\documentclass[runningheads,a4paper,oribibl]{llncs}

\usepackage{amssymb}
\usepackage{graphicx}

\usepackage{cite}
\usepackage{stfloats}
\usepackage{array}
\usepackage{multirow}
\usepackage{paralist}

\usepackage{url}
\urldef{\mailsa}\path|haobibo@gmail.com, lilindeqinchun@sina.com|
\urldef{\mailsb}\path|{gaorui11,liang08}@mails.ucas.ac.cn|
\urldef{\mailsc}\path|tszhu@psych.ac.cn|
\newcommand{\keywords}[1]{\par\addvspace\baselineskip
\noindent\keywordname\enspace\ignorespaces#1}

\begin{document}

\mainmatter  

\title{Sensing \emph{Subjective Well-being} from Social Media}

\titlerunning{Sensing SWB from Social Media}

%

%
\author{
Bibo Hao \inst{1} \and
Lin Li \inst{2} \and
Rui Gao \inst{1} \and
Ang Li \inst{1}\and
Tingshao Zhu \inst{1} \thanks{Corresponding author.}%
}
\authorrunning{B. Hao et al.}

\institute{ \{Institute of Psychology, University of Chinese Academy of Sciences\}, CAS \and
School of Humanities and Social Sciences, Nanyang Technological University\\
\mailsa\\
\mailsb\\
\mailsc\\
}

%
%

\toctitle{Sensing \emph{Subjective Well-being} from Social Media}
\tocauthor{Bibo Hao, Lin Li, Rui Gao, Ang Li and Tingshao Zhu}
\maketitle

\begin{abstract}
Subjective Well-being(\emph{SWB}),
which refers to how people experience the quality of their lives,
is of great use to public policy-makers as well as economic, sociological research, etc.
Traditionally,
the measurement of SWB relies on time-consuming and costly self-report questionnaires.
Nowadays, people are motivated to share their experiences and feelings on social media,
so we propose to sense SWB from the vast user generated data on social media.
By utilizing 1785 users' social media data with SWB labels,
we train machine learning models that are able to ``sense'' individual SWB.
Our model, which attains the state-of-the-art prediction accuracy,
can then be applied to identify large amount of social media users' SWB in time with low cost.


\end{abstract}

The last decade has witnessed the explosion of social media,
on which users generate huge volume of content every day.
Because of its richness and availability,
a lot of innovative research has been conducted on large scale social media data
to discover patterns in sociology, economics, psychology etc.,
which provides a brand new way for conventional social science research.
Studies have shown that
people's personal traits and psychological features,
such as gender, sexual orientation, personality, Intelligence Quotient and so on,
can be automatically predicted through clues on social media,
such as behavioral\cite{michal:pnas,plos14bfi} and linguistic\cite{upenn:language} patterns.

People pursue ``good life'' from ancient time to now,
and the Quality of Life (QoL)
is influenced by objective factors like income, jobs, health, environment,
which can be measured directly with objective indicators like GDP
or PM2.5 \footnote{Atmospheric Particulate Matter with diameter of 2.5 micrometers or less, which is an indicator of air pollution.}.
However, these objective factors cannot determine one's QoL.
The key indicator of QoL is Subjective Well Being (\textbf{SWB}),
encompassing emotional well-being and positive functioning\cite{corey:swb},
which refers to how people experience the quality of their lives and includes both emotional reactions and cognitive judgments.

Reliable and timely information of SWB,
provides important intellectual opportunities to research scientists and policy-makers. 
By analyzing large population data, it will be possible to identify the trend of SWB within different groups,
and figure out why some people are happy and others are not.
Many governments and organizations,
such as U.S.A, France,
OECD (Organization for Economic Co-operation and Development) etc.,
have been funding surveys and research to collect people's SWB data regularly
in order to support efficient decision making\cite{scimeg10objective,france10,oecd:swb}
and furthermore improve people's well-being.

Self-report survey is the conventional method which has been widely used to assess SWB.
Surveys comprise questions such as: How happy are you with your life?
Respondents answer a numerical scores (e.g., from very satisfied to very dissatisfied) in response to survey.
However, questionnaire surveys,
no matter in the form of paper-and-pencil, on-line etc.,
are costly and time consuming.
What's more, due to stereotype and social desirability,
participants may not provide accurate, honest answers
since survey is conducted in an intrusive manner -- asking questions to subjects.
Besides,
it is a big challenge to conduct questionnaire based surveys in large scale or carry out longitudinal study.

Recent studies focus on the prediction of psychological variables,
and the predicting models are established by analyzing the features and patterns of social media users' profiles, posts, likes, friends etc.
Such methods have been applied to the prediction of personality, depression, etc.
SWB prediction also attracts researches' attention\cite{schwartz2013ls},
while current works on SWB prediction are limited to prediction of groups other than individuals.
Some of these work even require costly census data like ``income median''.
Therefore, our goal in this work is to establish efficient SWB prediction model based on social media data,
which is applicable for individuals.

\section{Related Work}
\label{sec:related-work}

In this section, we will review foundations and studies related to our work in fields of psychology and computer science.

\vspace{1em}

\textbf{SWB and its Assessment}

Different from mere sentiment or simply happiness -- spontaneous reflections of immediate experience,
SWB is a measurement of individual's cognitive and affective evaluations of one's own life experience.
The structure of SWB we used in this paper, as listed in Table \ref{tab:swb},
is composed of \textbf{emotional well-being} and \textbf{positive functioning}.
Emotional well-being represents a long-term assessment towards one's life, which consists of two dimensions.
Positive functioning includes multidimensional structure of psychological and social well-being, and
psychological well-being encompasses six dimensions focusing on individual level.

Watson, Ryff et al., developed positive and negative affective scale (\textbf{PANAS})\cite{watson:panas}
and psychological well-being scale (\textbf{PWBS})\cite{ryff:pwb},
which are correspondent to emotional well-being and positive functioning respectively.
The reliability and validity of PANAS and PWBS
have been validated in long-term practices by numerous psychological studies.
In this paper, we use these two scales for SWB assessment.

\begin{table*}
\vspace{-0.5em}
\footnotesize
\def\firstcolwidth{47pt}

\centering
 \caption{Dimensions of Subjective Well-being and their description.}
    \label{tab:swb}
    \vspace{-1em}
\begin{tabular}{|>{\centering\arraybackslash}m{\firstcolwidth}|>{\centering\arraybackslash}m{59pt}|m{231pt}|}
\hline
        \multicolumn{2}{|l|}{\hfil{Dimension}} & \hfil{Description} \\ \hline
\multirow{2}{\firstcolwidth}{\hspace{-0.3em} \textbf{Emotional well-being} }
        & \texttt{Positive Affect} &\textbf{P.A.}  Experience symptoms that suggest enthusiasm, joy, and happiness for life. \vspace{0.3em} \\ \cline{2-3}
        & \texttt{Negative Affect} & \textbf{N.A.}  Experience symptoms that suggest that life is undesirable and unpleasant. \vspace{0.3em} \\ \hline
\multirow{5}{\firstcolwidth}{\textbf{Positive functioning}}
        & \texttt{Self Acceptance} & \textbf{S.A.}  Possess positive attitude toward the self; acknowledge and accept multiple aspects of self; feel positive about past life. \\ \cline{2-3}
        & \texttt{Purpose in Life} & \textbf{P.L.}  Have goals and a sense of direction in life; past life is meaningful; hold beliefs that give purpose to life. \\ \cline{2-3}
        & \texttt{Environmental Mastery} & \textbf{E.M.}  Feel competent and able to manage a complex environment; choose or create personally-suitable community. \\ \cline{2-3}
        & \texttt{Positive Relations with others} & \textbf{P.R.}  Have warm, satisfying, trusting relationships; are concerned about others’ welfare; capable of strong empathy, affection, and intimacy; understand give-and-take of human relationships. \\ \cline{2-3}
        & \texttt{Personal Growth} & \textbf{P.G.}  Have feelings of continued development and potential and are open to new experience; feel increasingly knowledgeable and effective. \\ \cline{2-3}
        & \texttt{Autonomy Items} & \textbf{A.I.} Are self-determining, independent, and regulate internally; resist social pressures to think and act in certain ways; evaluate self by personal standards. \\ \hline
\end{tabular}
\vspace{-1.5em}
\end{table*}

\textbf{Affect and Life Satisfaction Metric on Social Media}

Affect and Life Satisfaction (\textbf{LS}) reflect ``happiness'',
there are also recent studies investigated large scale social media data to metric people's affect or LS.

Quite a lot of studies use LIWC (Linguistic Inquiry and Word Count)\cite{pennebaker:liwc},
fruit carefully constructed over two decades of human research,
or other similar psychological language analysis tool,
to quantify psychological expression on social media. 
Representative works,
like hedonometer (happiness indicator) through Twitter by Dodds et al.\cite{sns-happiness},
twitter sentiment modeling and prediction of stock market by Bollen et al.\cite{bollen:twitter},
identify the sentiment (moods, emotions) in real time.
Furthermore, as ``face validation'',
the quantified metric is highly correlated to social events or economic indicators,
and it can even be predictable to economic trends.
By modeling people's sentiment through statuses and posts on SNS,
these works demonstrate that it is applicable to sense sentiment from social media.

It is noticeable that
recent works have introduced psychological as an assessment instrument for ``convergent validation''.
Convergent validity represents
to what degree a metric yielded by the model is similar to a psychological questionnaire based assessment.
Kramer proposed a model of ``Gross National Happiness''\cite{kramer:gnh} to predict satisfactory,
and used Diener and colleagues' SWL scale\cite{diener:swls} for convergent validation.
Please note that happiness defined in this work, is actually satisfaction to one's own life.

Burke and collages explored the relationship between particular activities on SNS and feelings of social capital\cite{sigchi:socialwb}.
They used Facebook Intensity Scale and UCLA loneliness scale to assess one's social capital feeling,
which could be used as an evaluation of one's cognitive feeling towards getting along with others.
However, on-line social well-being cannot cover the conception of SWB.

\vspace{0.5em}
\textbf{Predict Personal Traits and Mental Status via Social Media}

Kosinskia\cite{michal:pnas}, Schwartz\cite{upenn:language} et al.
analyzed the correlation between users' personal traits and behaviors or language usage on Facebook.
in which users' personality traits are measured by using Big-Five Personality Inventory.
They also build models to predict users' traits like personality through social media,
which is also another evidence of ``convergent validation'' approach.
Li et al.\cite{plos14bfi} use 839 behavioral features on microblog to predict personality,
which proved the feasibility of predicting psychological variables through behavioral features.
Similar works \cite{quercia2012personality} are also conducted 
on social media like Twitter.
Studies have also cast interests to the prediction of  mental health status via social media.
Hao\cite{hcii13mental}, Choudhury\cite{icwsm13-depression} et al. generalize this method to prediction depression, anxiety, etc.
analyzed users' both behavioral and linguistic features on microblog,
and employ machine learning methods to
predict depression, anxiety and other mental health status of individuals.

These pioneering works provide an innovative insight --
to predict (or in another word -- ``sense'') on-line users' psychological traits
or mental health status through his/her social media records.
Personality keeps relatively stable in one's life,
so accumulated on-line records can be enhanced evidence to predict one's personality.
While similar to mental health status, SWB is a psychological variable that varies over time.
Hence, predicting one's SWB should be based on one's behaviors in a specific period of time.

\vspace{0.5em}
\textbf{Prediction of Group SWB through Social Media}

Most recent work of Schwartz, Eichstaedt et al. generalize their method to \hbox{LS} prediction\cite{schwartz2013ls}.
Their work used LS as a single indicator of SWB,
and established model to predict the LS of each counties in the U.S.A through Twitter data.
In their work, county is the unit to predict the LS, rather than individual.
Their method, mainly analyze linguistic features on social media.
Furthermore, their model introduced variables like ``median age'', ``median household income'' and ``educational attainment'', which can only be obtained via costly census.

In this work, our goal is to establish model which can predict multi-dimension SWB of individuals,
considering both linguistic and behavioral features on social media.
Model based on individual will provide better generalization ability to different groups,
like groups with different ages, jobs and so on.

\section{Method}
\label{sec:method}
In our study,
we use both behavioral patterns and linguistic usage on social media, to identify their correlation with SWB.
In order to establish models,
we conduct a user study to collect user's social media data and SWB assessment.
Then, we treat the modeling problem as a typical machine learning problem:
to learn prediction model from social media data in which SWB is the label.

\subsection{Data Collection}
We ran our experiment on Sina Weibo (\url{http://weibo.com}),
a Chinese leading social media platform with over 300 million users
where more than 100 million microblogs are posted or reposted (retweeted) every day.

In the October of 2012,
we randomly sent inviting messages to about twenty thousand Weibo users who fulfill our requirements of ``active''.
Active users are defined as users who have posted more than 500 microblogs before recruiting.
Such active users have a relatively long term usage of social media,
and their Weibo statuses provide adequate information for analysis.

\begin{figure}
    \centering
    \includegraphics[width=0.8\textwidth]{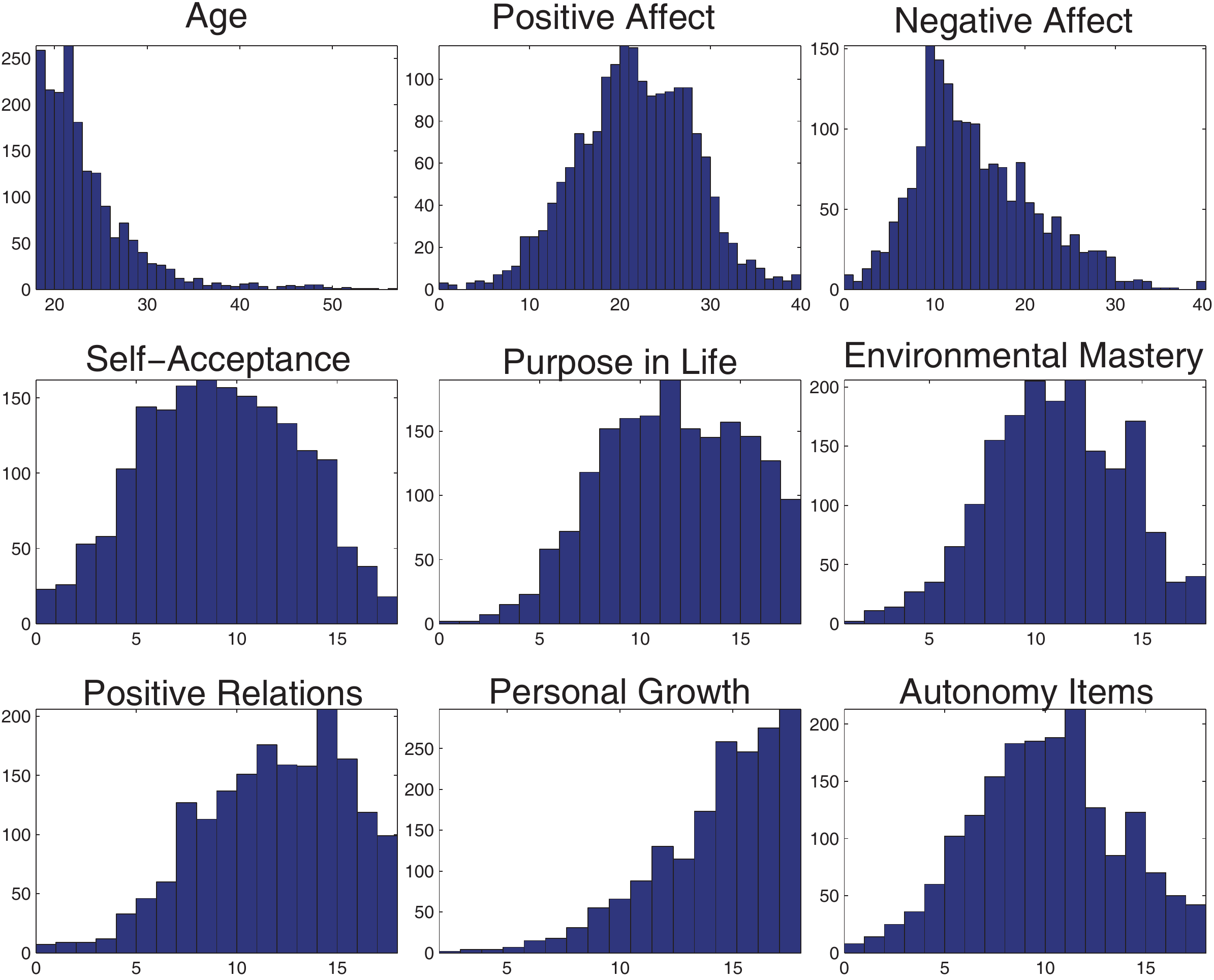}
    \caption{Distribution of Age and SWB Dimensions, where X-axis represents age or the score of each SWB dimension and Y-axis represents the number of participants. }
    \label{fig:hist}
    \vspace{-0.5em}
\end{figure}

Users who were willing to participate our experiment are guided to a web APP (\url{http://ccpl.psych.ac.cn:10002}).
Participants were then guided to agree an informed consent and fill psychological questionnaire.
Finally, 1785 adult volunteers (female:1136) filled the PANAS and PWBS survey
to assess their Emotional Well-being and Positive Functioning as SWB,
and their social media data were all downloaded through Sina Weibo API
one month after the user study.
Figure \ref{fig:hist} illustrates the distribution of participants' age and SWB distribution.
Our dataset contains users' social media data and SWB score.

\subsection{Experiment Design}

In previous studies,
researchers proposed different indicators (indexes) based on social media data,
and use psychological assessment as verification.
These indexes are defined subjectively,
and the procedure is always guided by researches' intuition of what factors in social media data may be correlated to target variable.
Actually, it is possible to build a prediction model by applying machine learning methods.
In this paper,
we take the task of predicting users' psychological variable based on their social media behaviors as a typical machine learning problem.
To do so, we extract features from users' social media data,
and train a machine learning model to predict the target variable (i.e., users' psychological variable).

\vspace{-1em}
\subsubsection{Golden Standard}
Core part of SWB sensing system
is the procedure of ``learning'' patterns of SWB from social media behavioral features.
To evaluate reliability of established model,
we use Pearson's correlation coefficient. 
In social psychology,
Pearson's correlation coefficient is a well-recognized measurement for convergent validity,
which is used to compare the relevance
between two assessing instruments or methods\cite{brackett2003convergent, duckworth2011meta}.
Specifically,
we calculate the correlation coefficient between psychological scales assessed SWB $\bf{Y}$,
and SWB sensed by a predicting model $\bf{\hat{Y}}$.
\vspace{-1.2em}
$$ \gamma = \rho ( \hat{\mathbf{Y}} , \mathbf{Y} ) = \frac{{\bf Cov}(\hat{\mathbf{Y}},\mathbf{Y})}{\sqrt{{\bf Var}(\hat{\mathbf{Y}}){\bf Var}(\mathbf{Y})}}. $$
\vspace{-1.1em}
\\The higher a model's $\bf{\gamma}$ is,
the closer the model can reach original scale.
In the work of Schwartz, Eichstaedt et al.\cite{schwartz2013ls}, they also adopted the same standard.

When measuring a psychological variable with different assessment instruments or methods,
correlation coefficient between different instruments or methods,
is typically around 0.39 to 0.68\cite{assess-personality}, i.e.: $\gamma \in [0.39, 0.68]$.
As a comparison, random guess (uniform distribution) yields $\gamma \in [-0.05, 0.05]$.

\vspace{-1em}
\subsubsection{Feature Extraction}
Our assumption in this study is that
one's SWB has impacts on one's behavioral or linguistic patterns on social media.
To predict SWB,
we adopted demographic, behavioral and linguistic features to build the predicting model.

\vspace{0.5em}

\textbf{Demographic Features (D, 3 features).}
In our case,
we use  \emph{\textbf{gender}},  \emph{\textbf{age}},
and  \emph{\textbf{category of living place}}\footnote{
We categorize living place in mainland China to
\begin{inparaenum}[a)]
  \item First-tier Cities: provincial capital and municipality cites, sample size $N_{3}=1009$;
  \item Other cities, $N_{2}=650$;
  \item Rural areas, $N_{1}=126$.
\end{inparaenum}
When using this features for regression, we simply let: ($LivingPlace=3,2,1$) respectively,
which can be seen as an indicator of population density.
Similarly, gender are set to 1 (male) and 0 (female).
}
categorized by population density as demographic features.
Although other demographic information, like ``educational attainment'', can be quite useful for SWB prediction,
they are actually unavailable on Weibo or many other social media platform.
The three features we extracted from social media profile are available in users' profile on most social media platform.

Notably, in our dataset,
when applying Student's t-test, we find that:
\begin{itemize}
  \item Except for \texttt{Negative Affect}, people live in first-tier cities, score significantly higher that people live in other areas in 7 dimensions of SWB ($p<0.005$);
  \item For \texttt{Positive Affect} and \texttt{Autonomy Items}, male users score significantly higher than female users ($p<0.005$);
  \item For \texttt{Negative Affect}, male users score significantly lower ($p<0.01$);
  \item Slight correlation occurs between users' age and SWB dimensions: \texttt{Environmental Mastery} 0.15, \texttt{Autonomy Items} 0.15 and \texttt{Negative Affect} -0.11.
\end{itemize}

These founding have also been reported by previous research\cite{diener1999swb}.
As found in our dataset, in China, living in first-tier city seems to offer a ``happier'' life,
although it means to be in a more competitive environment.
This might be caused by a comprehensive effect of income, education etc.

\vspace{0.5em}
\textbf{Behavioral Features (B, 26 features).}
We extract behavioral features from user profile and microblogs, including:
\vspace{-0.6em}
\begin{itemize}
  \item Interaction with other users, like following, friends and bi-following count;
  \item Express patterns, like microblog count, repost ratio within all statuses;
  \item Privacy protection, like whether enable geographical information, whether allowing ``strangers'' to comment;
  \item Personalization to social media access, like the length of nickname (on Weibo, users can change their nickname at any time).
\end{itemize}
\vspace{-0.2em}
These features generally describe users' implicit behavioral patterns on social media,
and they are available in user's detailed profile and microblog posts.

\vspace{0.5em}
\textbf{Linguistic Features (L, 88 features)}
SWB comprises abstract dimension like \texttt{Autonomy Items}, \texttt{Purpose in Life},
we believe such patterns might be implied in users' linguistic expression in microblogs.
Like many previous studies,
we use an improved version of LIWC, SCLIWC -- Simplified Chinese version LIWC optimized for microblog\cite{ccpl:cliwc},
to acquire users' linguistic patterns.
SCLIWC's dictionary categorizes words by psychological attributes, like Social Process, Percept, Personal Concern, etc.

\begin{table}[ht]
\vspace{-0.5em}
  \centering
    \caption{Correlation coefficient between 8 SWB dimensions and some features.}
    \label{tab:features}
    \vspace{-0.5em}
    \includegraphics[keepaspectratio=true, width=0.9\textwidth]{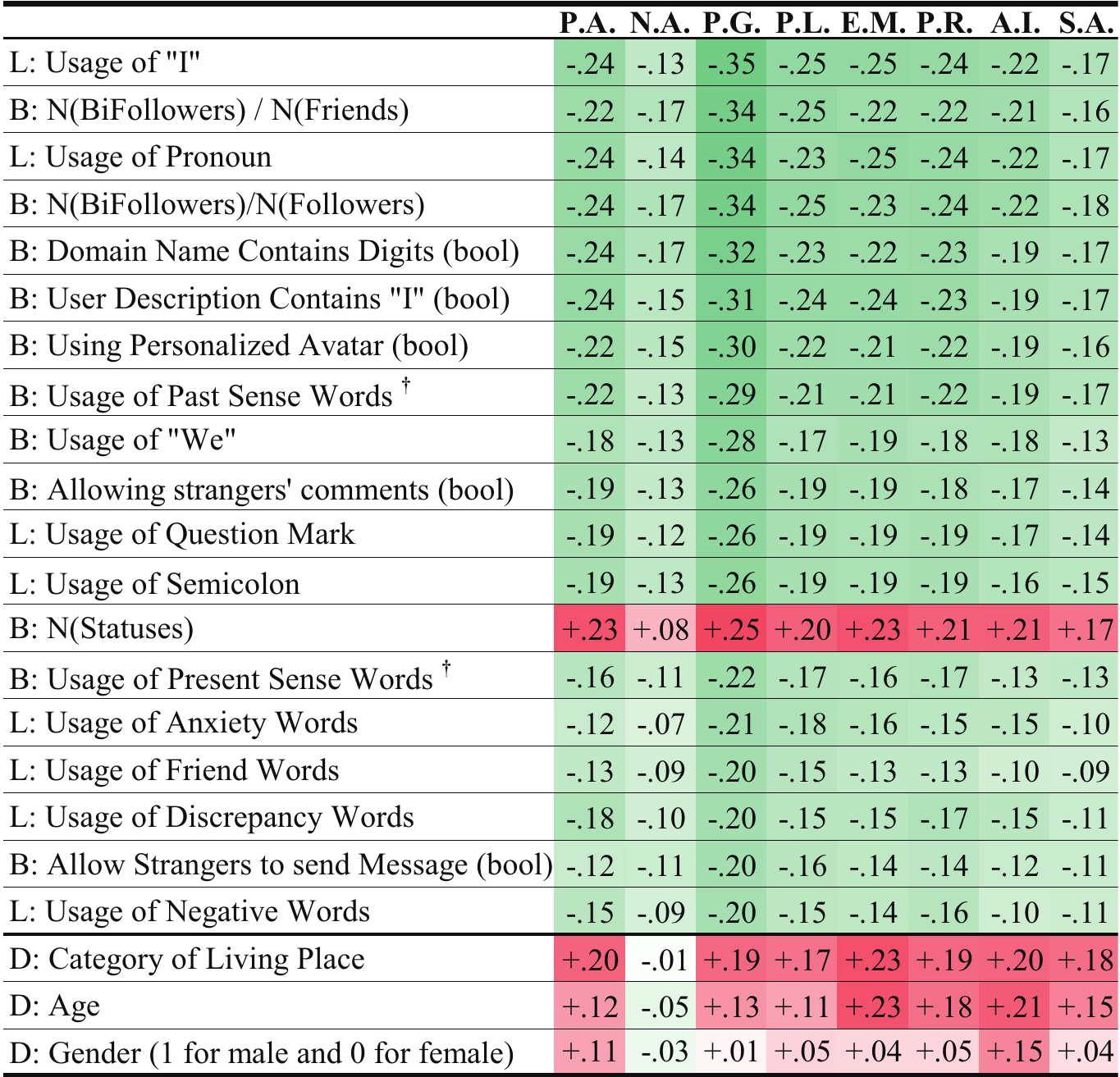}
    \leftline{ $^{\dag}$ Tense marking words are only available in Chinese.}
\vspace{-2em}
\end{table}

Since SWB may vary over time,
we extract linguistic features according to particular time period --
one week before and one week after the survey (denoted by $\pm 1Week$).
This is because our preliminary trial on different time point,
like 2 weeks before ($-2Week$),
2 weeks after ($+2Week$) filling questionnaire,
with simple linear regression algorithm reveals $\pm 1Week$ performs best.
We also compare the relevance of predicted SWB using 6 different combinations of feature set $\{D, B, L\}$,
and we take the regression model using only $\{D\}$ as baseline model.

\vspace{-1em}
\subsubsection{Feature Analysis}
Among all the 117 features,
we chose some features which are correlated with 8 SWB dimensions
in relatively high level and listed them in Table \ref{tab:features}.
The table shows the correlation coefficient between features and SWB dimensions,
from which we can see some behavioral and linguistic features on social media,
are positively or negatively correlated with SWB dimensions.
For example,
users using more first pronoun word ``I'' in language tend to have lower \texttt{Personal Growth};
users who posted more statuses on social media tend to have higher \texttt{Environmental Mastery}.
Such conclusions are in accordance with people's intuition, which can also be seen as a face validation of our method.

\vspace{-1em}
\subsubsection{Learning Algorithm}
SWB assessed by questionnaire survey comprises 8 dimensions, whose values are integers.
To build the SWB prediction model, our goal is,
for each SWB dimension, learn a function to maximize $\gamma$.

Since we didn't find algorithms targeting on maximizing Pearson's Correlation Coefficient,
we treated this problem as regression and tried following algorithms:

\vspace{-1em}
\begin{itemize}
	\item \textbf{Stepwise Regression}: Choose predictive variables Using F-test. 
	\item \textbf{LASSO} (Least Absolute Shrinkage and Selection Operator):
Using L1 norm to prevent overfitting, good at reducing feature space. 
	\item \textbf{MARS} (Multivariate Adaptive Regression Splines) non-parametric regression technique.
	\item \textbf{SVR} (Support Vector Regression) We used LibSVM implementation.

\end{itemize}

As SWB is widely used in social science and psychology,
although non-linear models may performs better,
it will be less interpretable that which factors impact SWB and how they contribute to SWB.
Whereas in linear models,
it is much easier to figure out which factors impact SWB, and their contributions.

Since the range of different features values are quite different,
data are normalized to keep features range in $[0,1]$:
$X_{i,j}' = ( X_{i,j} - minValue_{j} ) / ( maxValue_{j} - minValue_{j} )$.
Besides,
we apply \textbf{5-fold cross validation} to take most advantage of data and avoid potential overfitting on each algorithm.

\section{Results}
\label{sec:result}

As shown in Table \ref{tab:result},
we compare the performance of models trained by 4 algorithm on 6 combination of feature set.
In the left part, column \textbf{Feature Set} refers to feature combination,
for example, $B+D$ means to use Behavioral Features and Linguistic Features for training and testing.
Column \textbf{Algorithm} describes which algorithm is used to train learning model.
In the right part,
each cell shows the $\gamma$ value.
Darker cell background color means better performance.

At bottom of the table,
there are \textbf{``Feature Set Baseline''} -- models trained with only 3 demographic features.
Feature set baseline model perform poorly on each dimension, the $\gamma$ value is around $0.2$, which is a very weak correlation.
Additionally, performance in the case of random guess is basically $\gamma \in [-0.05, +0.05]$.

\begin{table}[Ht]
\vspace{-0.5em}
  \centering
    \caption{$\gamma$ Values: Pearson's Correlation Coefficient between SWB ``sensed'' by our model and assessed by PANAS/PWBS questionnaire scales.}
    \label{tab:result}
    \vspace{-1em}
    \includegraphics[keepaspectratio=true, width=0.7\textwidth]{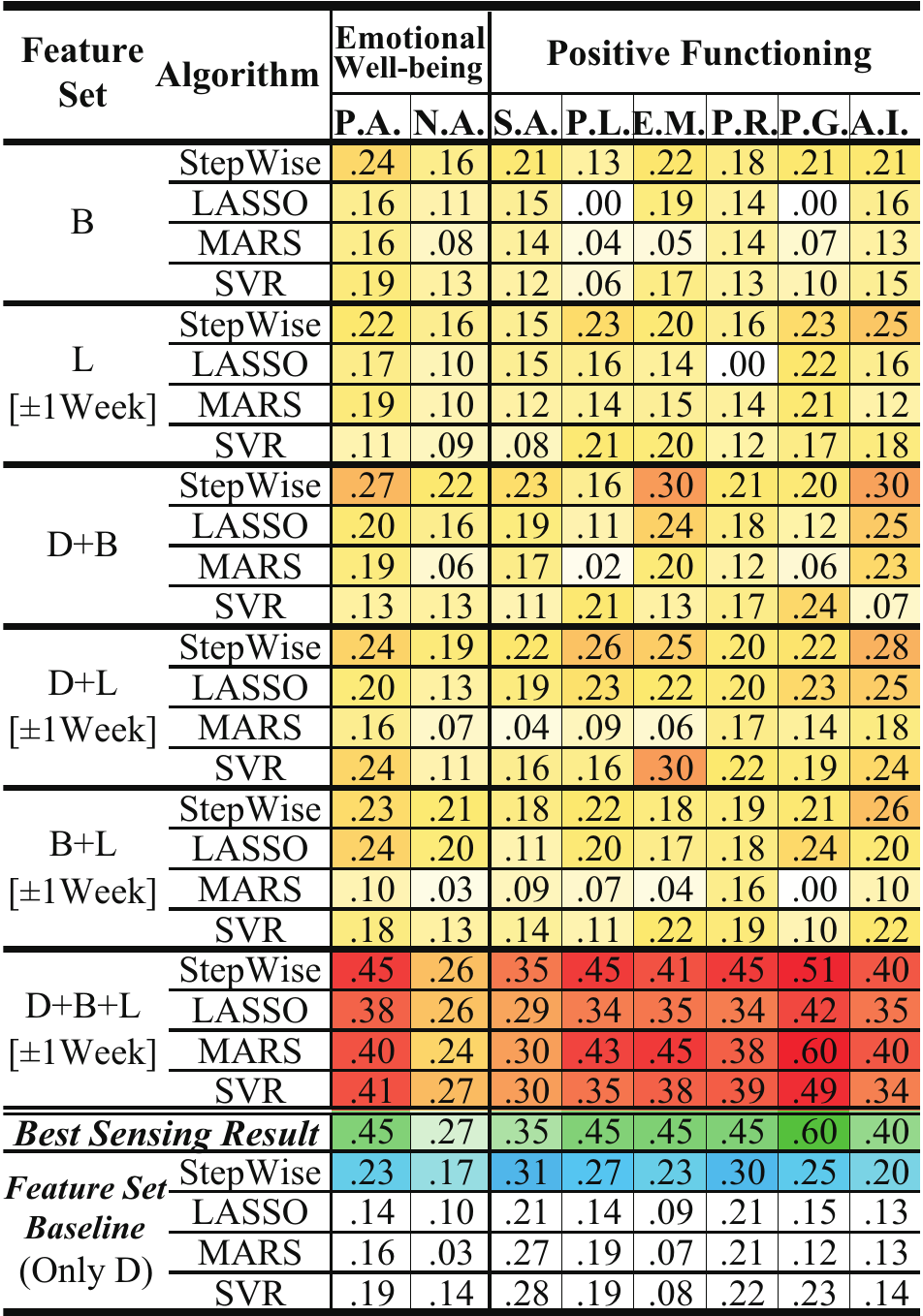}
\end{table}

Best performance of learning model is listed in the row of ``\textbf{Best Sensing Result}''.
It can be seen that our model performs fairly well in 7 dimension of SWB (except for \texttt{Negative Affect}).
As mentioned before, in social psychology research,
to particular psychological variable,
when a new developed assessing instrument or method achieves the standard of $\gamma \in [0.39, 0.68]$
with an existing reliable assessing method,
it is fair to say the new developed method has equivalent utility with the existing one.
As a comparison, work \cite{schwartz2013ls} predicts SWB of groups at the level of $\gamma \in [0.264, 0.535]$ using social media data,
and work \cite{scimeg10objective} predict SWB of groups at level of $\gamma=0.598$ using objective data.
Hence, our SWB prediction model has attained the state-of-the-art standard.

\section{Discussion}
\label{sec:discussion}

In our experiment, we tried 4 algorithms on 6 feature set combinations to establish models.
Experiment results show that feature set combination is significant for model performance.

\vspace{0.5em}
\textbf{Comparison of Features Set Combinations}
Models using only $B$, $L$ or $B+L$ actually  perform no better than baseline.
While adding demographic features into training feature set will improve the model performance to different extent.
Especially when we use all feature set ($D+B+L$), sensing model achieve best performance.
Although demographic feature set only contains age, gender and category of living place,
adopting these factors into feature set to train model will improve the model performance significantly.
This phenomenon also echoes to the work of \cite{schwartz2013ls},
after they added control factors like age, sex, monocytes, income and educational attainment into model,
prediction accuracy accrues from 0.307 to 0.535.

\vspace{0.5em}
\textbf{Comparison of Algorithms}
We adopted algorithms of linear and non-liner, parametric and non-parametric,
while in most cases, linear algorithm already perform fairly well.
In the same feature set of {$D+B+L$}, StepWise Regression performs better than other algorithms on 4 dimensions.
And MARS achieved $\gamma=0.6$ on dimension of Personal Growth, which is the best performance in all combinations.
Models trained using algorithm of LASSO contain relatively less features, for example, in the combination of {$D+B+L$},
44 features enter the final model to predict dimension of \texttt{Personal Growth}.

\vspace{0.5em}
\textbf{Limitation of This Work}
Like many other studies on social media,
our model also requires adequate user data for analysis.
In our experiment, we set the standard of ``active user'' as posting more than 500 microblog posts.
This limitation can be overcame along with users posting accumulating posts.


Weibo users accounts for more than a quarter of Chinese population,
which means, there are Weibo users in, if not every village, nearly every county.
SWB tendency of different groups can provide practical opportunities to policy-makers.
But social media users, surely cannot cover all population.

\section{Conclusions}
\label{sec:conclusion}

In this paper,
we established models to sense social media users' individual SWB, without survey or costly census data.
The established models, which attain the state-of-the-art prediction standard,
have equivalent utility with well-designed psychological scales.
This approach of psychological assessment,
can predict one's SWB by automatically by analyzing his/her social media data in a non-invasive manner,
and makes it feasible to assess users' psychological features, in large scale and timely.

Core of the paradigm in this study,
is to ``learn'' sensing (prediction) model from social media data and label data of psychological assessment.
Patterns and the interaction structures of explicit or implicit variables in social media,
can be automatically learned with algorithms (if they can be represented in the feature space).
Such paradigm avoids subjective bias of ``designing an index'' from numerous features,
in which case significant patterns may be hard to be discovered and adopted in the final model.
Besides, model is self-verified in the machine learning procedure using techniques like cross validation.


It is our will that the methods in this study can inspire subsequent research in the area of conventional psychology or social sciences.
More empirical analysis on real data, leads to more reliable conclusion, and such conclusion can be used to improve the public welfare.

\subsubsection*{Acknowledgements}
The authors gratefully acknowledges the generous support from
National High Technology Research and Development Program of China (2013AA01A606),
and Strategic Priority Research Program (XDA06030800).

\bibliographystyle{splncs}
\bibliography{SWB}

\end{document}